\documentclass[runningheads]{llncs}
\usepackage[T1]{fontenc}
\usepackage{amsmath}
\usepackage{amssymb}
\usepackage{dsfont}
\usepackage{tabularx}
\usepackage{graphicx}
\usepackage{amsmath}
\usepackage{algorithm}
\usepackage{subcaption}
\usepackage{algpseudocode}
\usepackage{lipsum} 
\usepackage{booktabs}
\usepackage{ulem}
\usepackage{arydshln}
\usepackage{multicol}
\usepackage{multirow}
\usepackage{hyperref}
\usepackage{tablefootnote} 
\begin{document}
%
\title{SOAEsV2-7B/72B: Full-Pipeline Optimization for State-Owned Enterprise LLMs via Continual Pre-Training, Domain-Progressive SFT and Distillation-Enhanced Speculative Decoding}
\titlerunning{SOAEsV2-7B/72B}
%
%
\author{%
Jingyang Deng\inst{1}$^{\dagger}$ \and
Ran Chen\inst{1}$^{\dagger}$ \and
Jo-Ku Cheng\inst{1} \and
Jinwen Ma\inst{1}$^{*}$
}

\renewcommand{\thefootnote}{} 
\footnotetext{$^{\dagger}$ These authors contributed equally to this work.}
\footnotetext{$^{*}$ Corresponding author.}
\renewcommand{\thefootnote}{\arabic{footnote}}

\institute{$^1$School of Mathematical Sciences and LMAM, Peking University, Beijing 100871, China \\
\email{jwma@math.pku.edu.cn}}
\maketitle              
\begin{abstract}
This study addresses key challenges in developing domain-specific large language models (LLMs) for Chinese state-owned assets and enterprises (SOAEs), where current approaches face three limitations: 1) constrained model capacity that limits knowledge integration and cross-task adaptability; 2) excessive reliance on domain-specific supervised fine-tuning (SFT) data, which neglects the broader applicability of general language patterns; and 3) inefficient inference acceleration for large models processing long contexts. In this work, we propose SOAEsV2-7B/72B, a specialized LLM series developed via a three-phase framework: 1) continual pre-training integrates domain knowledge while retaining base capabilities; 2) domain-progressive SFT employs curriculum-based learning strategy, transitioning from weakly relevant conversational data to expert-annotated SOAEs datasets to optimize domain-specific tasks; 3) distillation-enhanced speculative decoding accelerates inference via logit distillation between 72B target and 7B draft models, achieving 1.39–1.52$\times$ speedup without quality loss. Experimental results demonstrate that our domain-specific pre-training phase maintains 99.8\% of original general language capabilities while significantly improving domain performance, resulting in a 1.08$\times$ improvement in Rouge-1 score and a 1.17$\times$ enhancement in BLEU-4 score. Ablation studies further show that domain-progressive SFT outperforms single-stage training, achieving 1.02$\times$ improvement in Rouge-1 and 1.06$\times$ in BLEU-4. Our work introduces a comprehensive, full-pipeline approach for optimizing SOAEs LLMs, bridging the gap between general language capabilities and domain-specific expertise.
\end{abstract}

\section{Introduction}
\label{sec:introduction}

The integration of large language models (LLMs) into specialized industrial domains has emerged as a pivotal avenue for advancing digital transformation \cite{zhang2024cosis,shaikh2025fields}. Within the context of State-Owned Assets and Enterprises (SOAEs), intelligent solutions must satisfy three critical requirements: 1) accurate open-domain knowledge question-answering; 2) generation of industry-compliant professional reports; and 3) data-driven decision recommendations. These capabilities necessitate models that not only possess a deep understanding of industry-specific knowledge but also maintain general language abilities to address diverse demands. Current general-purpose models face a critical challenge in this domain—their overemphasis on generalizability leads to insufficient integration depth of SOAEs-specific expertise. This imbalance between specialization and generalization hampers models' ability to meet practical requirements for precise decision support and complex task processing \cite{song2025injectingdomainspecificknowledgelarge}, ultimately limiting the deep integration of AI technologies in industrial intelligence transformation.

Prior research in this field has made strides in domain-specific data curation \cite{huang2024soaes} and domain adaptation and preference alignment training strategies \cite{deng2025enhancing}. However, three systemic bottlenecks persist: First, existing SOAEs-specific LLMs are constrained by model capacity limitations, which hinder the integration of SOAEs' complex knowledge systems \cite{deng2025enhancing}. Second, current supervised fine-tuning methods typically rely on single-stage training with domain data \cite{huang2024soaes}, thereby neglecting the value of progressive knowledge transfer \cite{zhong2024seekingneuralnuggetsknowledge}, which enables models to gradually adapt to domain-specific tasks and accumulate expertise in a stepwise manner. Third, while current models with relatively smaller scales exhibit minimal inference bottlenecks, the efficiency of inference becomes a critical factor affecting practical usability as models scale up and support longer contexts \cite{zhou2024surveyefficientinferencelarge}.

In this study, we introduce the SOAEsV2-7B/72B model series, which addresses these challenges through the following approach: 1) continual pre-training on 72B-scale models robustly integrates domain knowledge and lays the groundwork for subsequent steps; 2) building on this foundation, a curriculum-driven, domain-progressive supervised fine-tuning (SFT) strategy constructs pathways for knowledge transfer from weakly relevant general data to strongly correlated domain-specific data; and 3) logit distillation-enhanced speculative decoding enables efficient cooperative inference between 7B and 72B models. This methodology provides SOAEs researchers with a comprehensive framework for full-pipeline optimization, spanning from incremental knowledge injection to efficient inference.

The key contributions of this work can be summarized as follows:

\begin{itemize}
    \item \textbf{Domain-Specific Model Scaling}: 
    Through continual pre-training, we have developed the largest 72B-parameter LLM in the SOAEs domain, thereby achieving significant advancements in both domain knowledge capacity and general capability compared to prior 7B models.
    
    \item \textbf{Domain-Progressive SFT}: 
    We introduce a curriculum-driven two-stage fine-tuning strategy. The first stage focuses on building essential cognitive abilities by utilizing general data that has been meticulously filtered for domain relevance. The second stage then delves into specialization, employing both expert-annotated data and data augmented by LLMs to further refine the model's domain expertise.
    
    \item \textbf{Cooperative Inference Acceleration}: 
    Our proposed method of logit-distilled speculative decoding facilitates decoding acceleration for the 72B target model with the assistance of a 7B draft model. This technique achieves a notable speedup of 1.39 to 1.52$\times$ without any sacrifice in accuracy.
    
    \item \textbf{Cross-Domain Methodological Insights}: 
    The design principles of our framework—staged training strategies and cooperative inference—offer a systematic methodology for LLM development in other domains. These insights may be applicable to other domains, such as finance \cite{li2023large} and healthcare \cite{yang2023large}, where domain knowledge and deployment efficiency are critical.
\end{itemize}

\section{Related Work}
\label{sec:related}

\subsection{SOAEs-Specific LLM Development}
Recent research on LLMs for SOAEs has advanced along two primary directions: domain-specific corpus development and specialized training approaches. However, despite these efforts, the research landscape in this domain remains underserved, with only a limited number of studies having explored this area \cite{huang2024soaes,deng2025enhancing}. Huang et al. \cite{huang2024soaes} pioneered this field by introducing SOAEs-DataSuite, which offered the first organized system for curating domain-specific data. Deng et al. \cite{deng2025enhancing} further enhanced this work by incorporating sequential training strategies that conduct domain adaptation followed by preference alignment. Despite these contributions, the existing research has predominantly centered on 7B-scale models and single-stage fine-tuning processes. Our research builds on this foundation by introducing several key innovations: 1) expanding model capacity to 72B parameters to facilitate more in-depth domain knowledge integration; 2) implementing curriculum-based Domain-Progressive SFT as a replacement for traditional single-stage SFT; and 3) developing cooperative inference mechanisms to support the deployment of large-scale models—areas that have not been adequately addressed in prior SOAEs-specific LLM research.

\subsection{Curriculum Learning in Domain Adaptation}
Curriculum learning \cite{bengio2009curriculum} has traditionally emphasized progressive task complexity or difficulty. In recent years, curriculum learning has been adopted in training LLMs and multi-modal LLMs, significantly enhancing performance across diverse domains, including code generation \cite{nair-etal-2024-curriculum}, speech processing \cite{hu-etal-2024-wavllm}, and even reinforcement learning for reasoning tasks \cite{liu2024letslearnstepstep}. Building on this foundation, our domain-progressive supervised fine-tuning (SFT) introduces a relevance-based curriculum that gradually shifts from weakly related general data to highly domain-specific content tailored for SOAEs. Unlike conventional curricula that are based on difficulty, our approach prioritizes the alignment between data relevance and the specific requirements of domain tasks. This staged training strategy allows for gradual specialization while maintaining general capabilities, which is essential for SOAEs applications that require a balance of domain expertise and open-domain proficiency. Our methodology thus presents a novel application of curriculum learning principles to the specialization of industrial LLMs.

\section{Methodology}
\label{sec:methodology}

\begin{figure}
\centering
\includegraphics[width=0.8\textwidth]{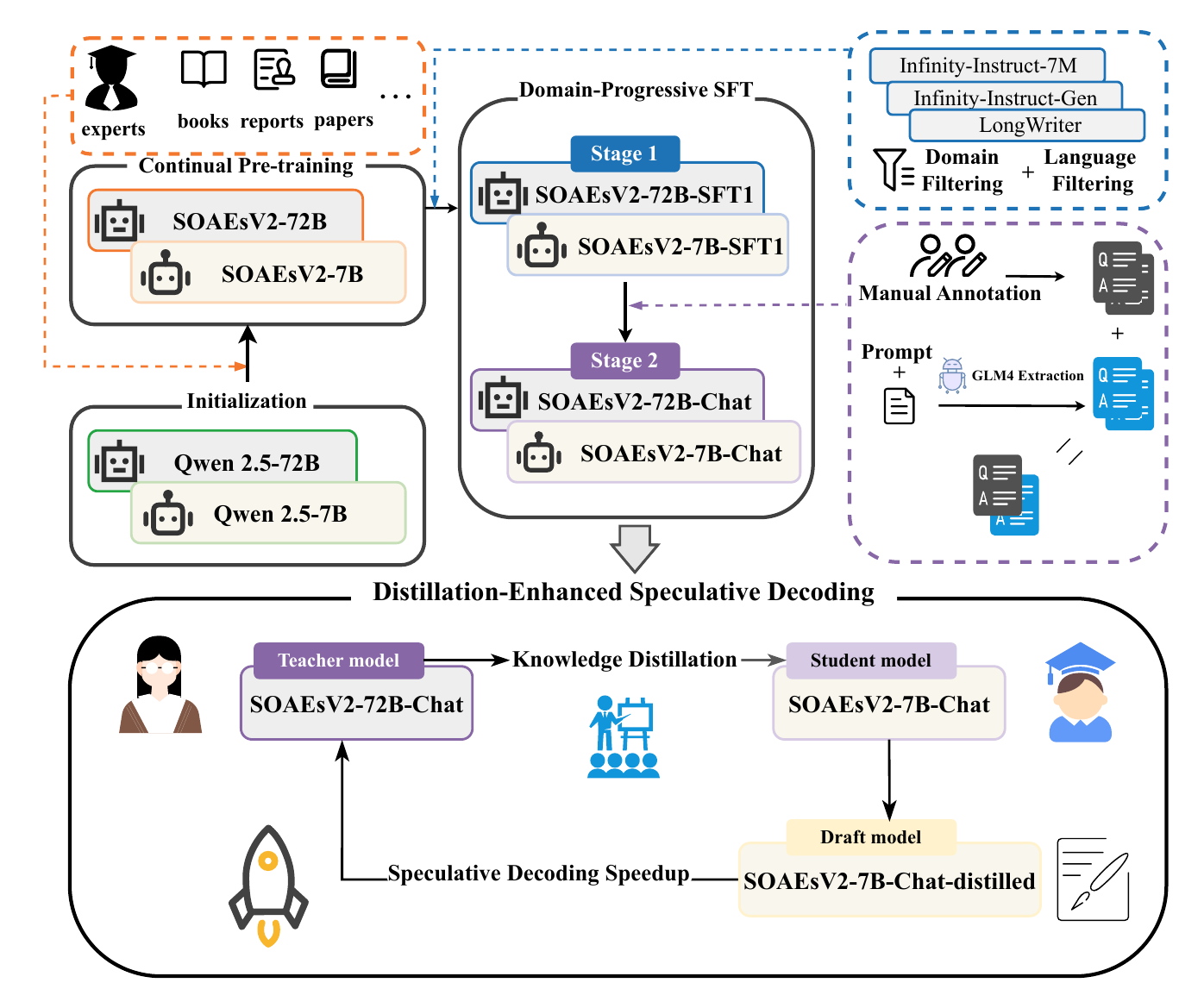}
\caption{Pipeline of our document layout analysis algorithm} \label{overview}
\end{figure}

In this section, we introduce our methodology for developing domain-specific Large Language Models (LLMs) in the SOAEs domain, as illustrated in Figure \ref{overview}. Our framework comprises three important phases: (1) \textbf{Continual Pre-training}, which preserves core capabilities while integrating domain knowledge; (2) \textbf{Domain-Progressive SFT}, implementing curriculum-based progression from general to specialized data; and (3) \textbf{Distillation-Enhanced Speculative Decoding}, facilitating efficient 7B-72B model coordination. 
The key insight of our method lies in the systematic optimization across the entire pipeline—from pre-training to inference. This approach first establishes foundational capabilities, then specializes through sequential training, and finally enables collaborative inference via aligned model distributions. This hierarchical approach ensures maximum knowledge retention while achieving both domain expertise and inference efficiency, as detailed in the following subsections.

\subsection{Continual Pre-Training}
\label{subsec:pretraining}

To address the challenge of integrating domain-specific knowledge, we conduct continual pre-training on a 72B-scale model. The primary goal of this phase is to inject domain knowledge into the model, while the secondary objective is to preserve its foundational language abilities. To efficiently reduce training costs, we meticulously curated a domain-specific corpus of 17B tokens from SOAEs-DataSuite \cite{huang2024soaes}. This was achieved through a rigorous expert-guided filtering process aimed at retaining the most essential and relevant subsets of SOAEs-DataSuite, ensuring a cost-effective and high-quality dataset for training. The filtering process prioritizes the credibility, relevance, and importance of the data sources. Additionally, to further ensure alignment with practical needs, we supplemented the corpus with additional data through consultations with domain experts. This combined approach ensures that the training data consists of knowledge-intensive, domain-relevant content that maximizes knowledge integration efficiency. The corpus was then preprocessed following the methodology outlined in \cite{huang2024soaes}, which included data parsing, deduplication, and format cleaning. This procedure minimizes data noise and ensures data quality, enabling effective knowledge injection into the model.

Unlike the approach in \cite{deng2025enhancing}, our work prioritizes allocating limited computational resources to larger-scale parameter models, focusing exclusively on domain-specific data without mixing extensive general-purpose corpora. The training strategy employs careful data curation to construct a relatively small dataset, combined with intentionally limited training iterations, thereby reducing the risk of catastrophic forgetting. The method ultimately achieves dual optimization—maintaining the model's general capabilities while enhancing domain-specific expertise, which is further corroborated by our experimental results in Section \ref{experiments}

\subsection{Domain-Progressive SFT}  
\label{subsec:domain_progressive_sft}  
Building upon the base model established during continual pre-training, our curriculum-driven domain-progressive SFT extends traditional curriculum learning principles through domain relevance stratification, progressively transitioning from weakly-related general data to core SOAEs expertise. This approach, unlike conventional single-stage SFT that struggles with abrupt domain shifts, constructs a sequential knowledge transfer pathway to mitigate this issue while optimizing domain adaptation effectiveness.

\textbf{Stage 1: Domain-Relevant Dialog Priming} 
establishes foundational competencies using conversational data filtered for latent SOAEs relevance, based on practical application requirements. We curate subsets from Infinity-Instruct-7M/Gen \cite{infinity-instruct} and LongWriter \cite{bai2024longwriter}, retaining domains with knowledge overlap in SOAEs contexts (finance, law, business, politics) while excluding unrelated categories (e.g., mathematics, arts). This phase focuses on question-response patterns, logical reasoning, and cross-domain information integration, with all data constrained to Chinese per application scenarios.

\textbf{Stage 2: Expert-Driven Specialization} 
focuses on enhancing domain expertise by leveraging human-annotated SOAEs data alongside synthetic data generated via a general-purpose LLM. In this stage, we have collected a dataset containing $\sim$33k samples, among which $\sim$1k are report generation tasks and the rest are short Q\&As. We randomly selected $\sim$80\% of the data as the training set and reserved the remaining data for subsequent development and testing. Specifically, to ensure the professional relevance of the generated Q\&A pairs, we prompt GLM4 \cite{glm2024chatglm} with domain-specific documents, guiding it to extract contextually relevant questions and answers. These synthetic Q\&A pairs are then subjected to rigorous human validation to ensure factual accuracy and alignment with domain-specific tasks. This process ensures that the generated content reflects authentic SOAEs application scenarios while maintaining high standards of domain-specific professionalism.

Our domain relevance stratification mechanism redefines curriculum learning by prioritizing task alignment over difficulty progression: In stage 1, weakly relevant data preserves general dialog capabilities while priming domain adaptation through related subdomains (e.g., financial analysis in SOAEs contexts), while in stage 2, strongly relevant, expert-verified content drives deep adaptation to SOAEs-specific patterns. This sequential strategy mirrors the natural progression of human expertise development—from broad industry literacy to specialized task mastery—providing a robust framework for progressive domain adaptation without overfitting to narrow patterns.

\subsection{Distillation-Enhanced Speculative Decoding}

To improve inference efficiency without sacrificing output quality, we introduce a distillation-enhanced speculative decoding (SPD) strategy. Following continual pre-training and SFT, we developed a 72B SOAEs domain-specific large model capable of handling diverse domain tasks. However, its inference speed posed significant bottlenecks. To resolve this, we train a 7B parameter draft model using identical data pipelines and combine logit-level distillation with an optimized draft-then-verify paradigm, establishing a principled end-to-end solution for accelerating domain-specialized LLM deployment.

\textbf{Logit Distillation for Alignment}.
To enable high-quality speculative decoding, the output distributions of the draft model \( \mathcal{M}_D \) and the target model \( \mathcal{M}_T \) must be closely aligned. We achieve this via logit-level distillation. For a given input \( x \), the objective is defined as:

\begin{equation}
\mathcal{L}_{\text{KL}} = \text{KL} \left( \text{softmax}\left( \frac{z_T(x)}{\tau} \right) \| \text{softmax}\left( \frac{z_D(x)}{\tau} \right) \right)
\end{equation}

\begin{equation}
\mathcal{L} = \alpha \cdot \mathcal{L}_{\text{KL}} + (1 - \alpha) \cdot \mathcal{L}_{\text{SFT}}, \quad \alpha \in [0, 1]
\end{equation}
where \( z_T(x) \) and \( z_D(x) \) denote the logits of the target and draft models respectively, \( \tau \) is a temperature parameter to soften predictions, and \( \alpha \) balances distillation with task-specific supervision. This process ensures the draft model can serve as a reliable proxy during decoding, maximizing token agreement and downstream acceptance.

\textbf{Speculative Decoding with Aligned Draft Model}.
Speculative decoding \cite{leviathan2023fast} provides a general framework for latency reduction by decoupling token generation and verification. Compared to prompt lookup \cite{saxena2023prompt} decoding which relies on heuristic n-gram matching, we leverage a dedicated distilled draft model that enables more accurate and efficient speculative predictions through logit-level distillation alignment. The acceleration comes from two key mechanisms: 1) the draft model's lightweight nature enables fast autoregressive proposal generation, and 2) the target model's parallel verification of multiple tokens amortizes its computational overhead. By having the powerful target model only validate batches of tokens rather than generate each token sequentially, the framework significantly reduces end-to-end latency while maintaining distributional equivalence.

The decoding workflow proceeds as follows:

\begin{enumerate}
    \item Draft Generation: The draft model \( \mathcal{M}_D \) autoregressively proposes a block of \( n \) tokens \( \hat{y}_{1:n} \) based on current context \( X \), where each \( \hat{y}_i \) is the token generated by the draft model at position \( i \).
    
    \item Parallel Verification: The target model \( \mathcal{M}_T \) evaluates the same context and returns token predictions \( y_{1:n} \), where each \( y_i \) is the token predicted by the target model at position \( i \). Each draft token is accepted if it matches:

    \begin{equation}
    \text{Accept}_i = \mathbb{I} \left[ \hat{y}_i = y_i \right], \quad \forall i \in \{1, \dots, n\}.
    \end{equation}

    \item Rejection Sampling (skipped when using greedy decoding): For probabilistic sampling decoding, let \( k \) denote the first rejected position. Tokens \( \hat{y}_{1:k-1} \) are retained, while the token at \( k \) is resampled using the residual distribution:

    \begin{equation}
    p_{\text{res}}(x) = \frac{\max \left( p_T(x) - p_D(x),\, 0 \right)}{\sum_{x'} \max \left( p_T(x') - p_D(x'),\, 0 \right)},
    \end{equation}

    where \( p_T(x) \) and \( p_D(x) \) represent the probability distributions predicted by the target model and draft model for token \( x \), respectively, and \( x' \) represents all possible tokens in the vocabulary. This ensures exact equivalence to the target sampling distribution. When using greedy sampling, this step is unnecessary as verification failures simply truncate the draft sequence.

    \item Iteration: The accepted prefix and the new token form the next decoding context. This procedure repeats until generation is complete.
\end{enumerate}
This framework transforms autoregressive decoding into a hybrid process where the target model only intervenes selectively, reducing its computational load. The effectiveness of speculative decoding relies heavily on the agreement between the draft and target distributions, which our distillation step explicitly optimizes.

\section{Experiments}

\subsection{Exprimental Settings}

All experiments were conducted on 8$\times$A800 80G GPUs with full-parameter training under an 8k context window length. Starting from the Qwen2.5-7B/72B \cite{yang2024qwen2} model as our foundation, we utilized DeepSpeed Zero-3 \cite{rasley2020deepspeed} with CPU offload, gradient checkpointing \cite{chen2016training}, and Flash Attention v2 \cite{dao2023flashattention} for training, with a warmup ratio of 0.05. For continual pre-training, we utilized an total batch size of 64, corresponding to $\sim$ 0.5M tokens, and a learning rate of \(1 \times 10^{-5}\), which was decayed to \(1 \times 10^{-6}\) using cosine scheduling. The first-stage SFT employed a packing \cite{krell2021efficient} strategy to concatenate multiple sequences within the 8k context window. We retained the total batch size of 64 and maintained a constant learning rate of \(1 \times 10^{-6}\) to stabilize foundational capability retention. The second-stage SFT similarly leveraged the packing strategy but reduced the batch size to 8 while increasing the learning rate to \(3.5 \times 10^{-6}\) to focus on domain adaptation. Logit distillation was conducted with \(\alpha = 0.5\), temperature \(\tau = 2.0\), batch size 8, and a learning rate of \(3.5 \times 10^{-6}\). We stored the logits of the 72B model offline for this process to facilitate training. Speculative decoding utilized a block size of $n=3$ and top-\(p = 0.7\) for candidate token generation. All training phases, including continual pre-training, SFT, and distillation, were executed for 1 epoch to avoid over-fitting.

\begin{table}[htbp]
\centering
\caption{Benchmark Evaluation of General Language Understanding Capabilities Before and After Continual Pre-training}
\label{table1}
\begin{tabular}{lccc}
\toprule
Model                 & CMMLU         & C-EVAL  & Avg.          \\ \midrule
SOAEsV1-7B          & 71.5          & 71.0                                         & 71.3          \\
SOAEsV2-7B (ours)      & 81.8          & 80.3                                       & 81.1          \\
Qwen2.5-7B       & 82.9          & 81.3                                        & 82.1          \\
 \midrule
DeepSeek-V3 base\tablefootnote{Results are from \cite{liu2024deepseek}.} & 88.8          & \uline{90.1}                       & 89.5       \\
SOAEsV2-72B (ours)      & \uline{90.2} & 90.0                                        & \uline{90.1}        \\
Qwen2.5-72B      & \textbf{90.4}          & \textbf{90.2}                                         & \textbf{90.3}          \\
 \bottomrule
\end{tabular}
\end{table}

\subsection{Main Resutls}
\label{experiments}

\textbf{Retained General Capabilities via Continual Pre-training}. Table~\ref{table1} evaluates the retention of general Chinese capabilities after continual pre-training using the CMMLU \cite{li2023cmmlu} and C-EVAL \cite{huang2023c} benchmarks. Three key findings include: 1) SOAEsV2-72B retains 99.8\% of the base performance of Qwen2.5-72B (90.1 vs. 90.3 avg.), outperforming the 671B DeepSeek-V3 base \cite{liu2024deepseek} Mixture-of-Experts model by 0.6 points, indicating that large-scale domain specialization does not significantly impair generalization. 2) SOAEsV2-7B achieves a 13.7\% improvement over its predecessor SOAEsV1-7B (from \cite{deng2025enhancing}, 81.1 vs. 71.3 avg.), narrowing the performance gap to Qwen2.5-7B to 98.8\% (81.1 vs. 82.1 avg.), primarily attributed to leveraging a stronger base model (Qwen2.5-7B) for continual pre-training. 3) Catastrophic forgetting diminishes with model scaling: SOAEsV2-72B retains 99.8\% of its base capabilities compared to 98.8\% for the 7B variant, underscoring the critical role of scaling laws in preserving knowledge. These results collectively demonstrate the framework's ability to balance domain expertise with general capabilities, while subsequent analyses validate the complementary benefits of domain-progressive SFT and distillation-enhanced inference.

\begin{table}[htbp]
\centering
\caption{Comparative Analysis of SOAEsV2-72B Models Across Training Stages}
\label{table2}
\begin{tabular}{lcccc}
\toprule
Model                                          & Rouge-1        & \multicolumn{1}{l}{Rouge-2} & \multicolumn{1}{l}{Rouge-l} & BLEU-4         \\ \midrule
Qwen2.5-72B-Instruct                           & 28.26          & 9.37                       & 17.74                       & 10.69          \\
Qwen2.5-72B + stage-1 SFT              & 31.40          & 12.84                       & 21.03                       & 12.86          \\
\midrule
Qwen2.5-72B + stage-2 SFT                & 38.87          & 21.05                       & 29.90                       & 20.20          \\ 
Qwen2.5-72B + CPT + stage-2 SFT             & \uline{42.14}    & \uline{25.19}                 & \uline{34.04}                 & \uline{23.61}    \\ \midrule
Qwen2.5-72B-Instruct + stage-2 SFT & 39.59         & 20.80                       & 30.25                       & 20.05          \\ 

Qwen2.5-72B + CPT + stage-1\&2 joint SFT      & 41.08          & 23.95                       & 32.66                       & 22.63          \\
SOAEsV2-72B-Chat\tablefootnote{SOAEsV2-72B-Chat: Qwen2.5-72B + CPT + stage-1 SFT + stage-2 SFT} (ours) & \textbf{43.08} & \textbf{26.11}              & \textbf{35.11}              & \textbf{25.11} \\ \bottomrule
\end{tabular}

\end{table}

\textbf{Continual Pre-training Establishes Foundational Domain Expertise.} Our empirical analysis demonstrates that continual pre-training (CPT) is critical for grounding models in domain-specific knowledge. As shown in Table~\ref{table2}, models with CPT consistently outperform their non-CPT counterparts when subjected to identical stages of supervised fine-tuning (SFT). For instance, the Qwen2.5-72B model with CPT and stage-2 SFT achieves a Rouge-1 score of 42.14 and a BLEU-4 score of 23.61, representing an 8.4\% improvement in Rouge-1 and a 16.9\% improvement in BLEU-4 over the non-CPT baseline (38.87/20.20). Notably, models without CPT exhibit persistent performance gaps even after extensive domain-specific SFT, underscoring the necessity of CPT as a foundational step for effective domain adaptation.

\textbf{Domain-Progressive Staging Optimizes SFT.} In Table~\ref{table2}, our investigation reveals that curriculum-based progressive staging synergizes with CPT to optimize domain-specific downstream tasks. Comparative analysis shows that the CPT + stage-1 + stage-2 SFT configuration (43.08/25.11) achieves incremental gains of 2.2\% in Rouge-1 and 6.4\% in BLEU-4 over the CPT + stage-2-only SFT configuration (42.14/23.61). This supports our perspective that preliminary stage-1 SFT, leveraging domain-adjacent data, facilitates progressive specialization. Importantly, joint training of both stages (41.08/22.63) underperforms the staged approach by 4.6\% in Rouge-1 and 9.9\% in BLEU-4, indicating that indiscriminate data mixing disrupts coherent domain specialization. These findings establish domain-progressive SFT as a critical enabler of model performance rather than a trivial implementation detail. It is also worth noting that Qwen2.5-72B-Instruct \cite{yang2024qwen2}, while serving as a baseline in our experiments, is recognized as the state-of-the-art open-source general-purpose 72B model. Our SOAEsV2-72B-Chat further demonstrates superior effectiveness in domain adaptation by outperforming it.

\begin{table}[htbp]
\centering
\caption{Distillation Hyper-parameters Impact on Downstream Task Performance}
\label{table3}
\resizebox{\textwidth}{!}{%
\begin{tabular}{lcccccc}
\toprule
\multirow{2}{*}{Model} & \multicolumn{2}{c}{Distillation Parameters} & \multicolumn{4}{c}{Evaluation Metrics} \\ \cmidrule(lr){2-3} \cmidrule(lr){4-7}
 & $\alpha$ & $\tau$ & Rouge-1 & Rouge-2 & Rouge-l & BLEU-4 \\ \midrule
\multicolumn{7}{l}{\textbf{Undistilled Models}} \\ \midrule
\multirow{1}{*}{Qwen2.5-7B-Instruct} & - & - & 27.18 & 8.51 & 17.37 & 9.90 \\
\multirow{1}{*}{SOAEsV2-7B-Chat (undistilled)} & - & - & 34.11 & 15.81 & 26.06 & 15.61 \\ \midrule
\multicolumn{7}{l}{\textbf{Distilled Models with Varying $\alpha$ (Fixed $\tau=2.0$)}} \\ \midrule
\multirow{1}{*}{SOAEsV2-7B-Chat (distilled)} & 0.2 & 2.0 & 36.13 & 17.09 & 27.47 & 16.68 \\
\multirow{1}{*}{SOAEsV2-7B-Chat (distilled, ours)} & 0.5 & 2.0 & \textbf{36.37} & \textbf{17.28} & \textbf{27.82} & \textbf{17.37} \\
\multirow{1}{*}{SOAEsV2-7B-Chat (distilled)} & 0.8 & 2.0 & 34.93 & 16.08 & 26.73 & 15.69 \\
\multirow{1}{*}{SOAEsV2-7B-Chat (distilled)} & 1.0 & 2.0 & 33.63 & 15.03 & 24.94 & 13.80 \\ \midrule
\multicolumn{7}{l}{\textbf{Distilled Models with Varying $\tau$ (Fixed $\alpha=0.5$)}} \\ \midrule
\multirow{1}{*}{SOAEsV2-7B-Chat (distilled)} & 0.5 & 1.0 & 35.04 & 16.39 & 26.64 & 16.53 \\
\multirow{1}{*}{SOAEsV2-7B-Chat (distilled, ours)} & 0.5 & 2.0 & \textbf{36.37} & \textbf{17.28} & \textbf{27.82} & \textbf{17.37} \\
\multirow{1}{*}{SOAEsV2-7B-Chat (distilled)} & 0.5 & 3.0 & 35.50 & 16.56 & 27.18 & 16.13 \\
\multirow{1}{*}{SOAEsV2-7B-Chat (distilled)} & 0.5 & 5.0 & 35.08 & 16.12 & 26.87 & 16.04 \\ \bottomrule
\end{tabular}
}
\end{table}

\textbf{Optimal Distillation Parameters for Downstream Task Performance}. As evidenced in Table~\ref{table3}, our analysis reveals essential parameter interactions during knowledge distillation from the 72B teacher to the 7B student. The optimal balance occurs at balancing weight $\alpha=0.5$, where distillation loss and SFT loss reach equilibrium—higher $\alpha$ values (0.8–1.0) suppress the SFT loss weight ($1-\alpha$), inducing excessive teacher dependency that degrades generalization by 1.44 to 2.74 Rouge-1 score, while lower $\alpha$ (0.2) inadequately leverages teacher expertise. Simultaneously, temperature $\tau=2.0$ optimally adjust the teacher's distribution: excessive smoothing ($\tau\geq3.0$) dilutes task-specific knowledge in logits, whereas insufficient smoothing ($\tau\leq1.0$) creates overconfident distributions degrading to vanilla SFT, reducing Rouge-1 score by 0.87 to 1.33. This dual optimization enables our configuration ($\alpha=0.5$, $\tau=2.0$) to attain peak metrics (36.37 Rouge-1 and 17.37 BLEU-4).

\begin{table}[htbp]
\centering
\caption{Impact of Distillation Hyper-parameters on Speculative Decoding (\( n=3 \))}
\label{table4}
\resizebox{\textwidth}{!}{%
\begin{tabular}{lccccc}
\hline
Model & \(\alpha\) & \(\tau\) & Acceptance Rate (\%) & Speed (tokens/s) & Speedup \\
\hline
None & - & - & - & 28.95 & 1.00$\times$ \\
Qwen2.5-7B-Instruct & - & - & 68.06 & 36.07 & 1.25$\times$ \\
SOAEsV2-7B-Chat (undistilled) & - & - & 73.94 & 39.42 & 1.36$\times$ \\
SOAEsV2-7B-Chat (distilled) & 0.2 & 2 & 73.35 & 39.12 & 1.35$\times$ \\
SOAEsV2-7B-Chat (distilled, ours) & 0.5 & 2 & \textbf{75.78} & \textbf{40.26} & \textbf{1.39$\times$} \\
SOAEsV2-7B-Chat (distilled) & 0.8 & 2 & 74.40 & 39.40 & 1.36$\times$ \\
SOAEsV2-7B-Chat (distilled) & 0.5 & 1 & 73.99 & 39.49 & 1.36$\times$ \\
SOAEsV2-7B-Chat (distilled) & 0.5 & 3 & 73.45 & 39.28 & 1.36$\times$ \\
\hline
\end{tabular}
}
\end{table}

\textbf{Distillation-Enhanced Speculative Decoding Performance}.
Following distillation training, our framework enables speculative decoding by leveraging the 7B model as a draft model to accelerate the 72B target model. As shown in Table~\ref{table4}, the distilled SOAEsV2-7B-Chat draft model with optimal hyper-parameters (\(\alpha{=}0.5\), \(\tau{=}2\)) achieves a 1.39$\times$ speedup over the baseline SOAEsV2-72B-Chat with no acceleration, delivering 75.78\% acceptance rate and 40.26 tokens/s throughput (measured using the vLLM \cite{kwon2023efficient} framework under single-concurrency conditions). This configuration, previously validated for superior downstream task performance, outperforms alternative distillation settings, confirming that balanced weight \(\alpha\) and temperature \(\tau\) are critical for maximizing draft-target synergy. The results validate that our logit distillation strategy structurally optimizes the draft model’s output distribution for speculative decoding while preserving generation quality, enabling efficient acceleration of the 72B model without compromising accuracy.

\begin{table}[htbp]
\centering
\caption{Comparison of SPD and Prompt Lookup Decoding Speeds for Different n Values}
\label{table5}
\begin{tabular}{lccccc}
\hline
Decoding Method & n & Speed (tokens/s) & Speedup \\
\hline
Autoregression  & - & 28.95 & 1.00$\times$ \\
\hline
\multirow{2}{*}{SPD} & 3 & \textbf{40.26} & \textbf{1.39$\times$} \\
 & 5 & 39.42 & 1.36$\times$ \\
\hline
\multirow{2}{*}{Prompt Lookup} & 3 & 32.46 & 1.12$\times$ \\
 & 5 & 31.01 & 1.07$\times$ \\
\hline
\end{tabular}
\end{table}

\subsection{Ablation Studies}

\textbf{Ablation Study on Draft Model Necessity and Block Size Selection}. 
While existing acceleration methods like prompt lookup decoding \cite{saxena2023prompt} achieve draft-free inference through heuristic n-gram matching, our experiments demonstrate clear advantages of training dedicated draft models. As shown in Table~\ref{table5}, our SPD with distilled 7B draft model achieves 1.39$\times$ speedup ($n=3$), substantially outperforming prompt lookup's 1.12$\times$ under identical settings. This 24\% relative improvement validates that the logit-level distillation alignment between 72B target and 7B draft models enables more accurate speculative predictions than heuristic approaches. Furthermore, both methods exhibit decreasing speedups with larger block size ($n=3$ vs $n=5$), suggesting smaller blocks better balance verification overhead and speculative potential. Therefore, we choose $n=3$ in our main experiments.

\begin{table}[htbp]
\centering
\caption{Comparison of Greedy vs. Top-p Sampling Decoding Speeds}
\label{table6}
\begin{tabular}{lccccc}
\hline
Decoding Strategy & Decoding Method & Speed (tokens/s) & Speedup \\
\hline
\multirow{2}{*}{Top-p} & Autoregression & 28.95 & 1.00$\times$ \\
 & SPD & \textbf{40.26} & \textbf{1.39$\times$} \\
\hline
\multirow{2}{*}{Greedy} & Autoregression & 29.34 & 1.00$\times$ \\
 & SPD & \textbf{44.60} & \textbf{1.52$\times$} \\
\hline
\end{tabular}
\end{table}

\textbf{Decoding Strategy Analysis}. As shown in Table~\ref{table6}, while our primary application scenario requires controlled randomness via top-p sampling (yielding 1.39$\times$ speedup), greedy decoding achieves a maximal speedup of 1.52$\times$ (44.60 vs 29.34 tokens/s). The consistent acceleration across both strategies validates the efficiency of our distillation-enhanced speculative decoding framework.

\section{Conclusion}
\label{sec:conclusion}

In this work, we present SOAEsV2-7B/72B and establish a full-pipeline methodology for building LLMs tailored to the SOAEs domain. Our approach addresses critical limitations in prior domain-specific LLM development by introducing a three-phase optimization pipeline: continual pre-training, domain-progressive SFT, and distillation-enhanced speculative decoding. Through these efforts, we achieved significant improvements in domain-specific tasks while maintaining general-purpose language abilities. Future work could explore extending this methodology to other specialized domains such as finance and healthcare, where similar challenges exist. Our contributions not only advance SOAEs-specific AI solutions but also offer valuable insights for the broader LLM community.

\subsubsection{Acknowledgements}
This work was supported by the National Natural Science Foundation of China (62471175) and the high-performance computing platform of Peking University. 

\bibliographystyle{splncs03_unsrt}
\bibliography{references.bib}

\end{document}